\documentclass{article}
\usepackage{comment}
\usepackage{algorithm}
\usepackage{amsmath}
\usepackage{multirow}
\usepackage[utf8]{inputenc}
\usepackage{graphicx}
\usepackage[sorting=none]{biblatex}
\usepackage[a4paper,
            bindingoffset=0.2in,
            left=1in,
            right=1in,
            top=1in,
            bottom=1in,
            footskip=.25in]{geometry}
\usepackage{color,soul}
\usepackage{caption}
\usepackage{subcaption}
\usepackage{arydshln}
 \newcommand\blfootnote[1]{
    \begingroup
    \renewcommand\thefootnote{}\footnote{#1}
    \addtocounter{footnote}{-1}
    \endgroup
}
\bibliography{references.bib}

\title{Extending the Horizon of Early Diagnosis: Lung Cancer Prediction with Vision Transformers}

\author{Olivera Kotevska$^1$, Ian Goethert$^3$, Michael McGee$^3$, \\ Maria Mahbub$^4$,  Sean R. Wilkinson$^5$, Rowena Yip$^6$,  Myvizhi Esai Selvan$^6$, \\ Zeynep H. G\"um\"u\c s$^6$, Claudia Henschke$^6$, Robert J. Klein$^6$, Providencia Morales $^7$, \\ Samuel M Aguayo$^7$, Ioana Danciu$^{4,8}$, Mayanka Chandrashekar$^{2*}$}

\begin{document}
\date{}
 \maketitle
\parbox{\textwidth}{%
\centering
     $^1$ Mathematics in Computation, Computer Science and Mathematics Division,\\ Oak Ridge National Laboratory \\
     $^2$Advanced Computing for Health Sciences, Computational Sciences and Engineering Division,\\ Oak Ridge National Laboratory \\
     $^3$ Research Computing Support Division, \\ Oak Ridge National Laboratory \\
     $^4$ Advanced Intelligent Systems, Cyber Resilience and Intelligence Division, \\ Oak Ridge National Laboratory \\
     $^5$ National Center for Computational Sciences, Oak Ridge National Laboratory \\
     $^6$ Icahn School of Medicine at Mount Sinai \\
     $^7$ Phoenix VA Medical Center \\
     $^8$ Department of Biomedical Informatics, Vanderbilt University Medical Center \\
     $^*$ Corresponding Author Email: chandrashekm@ornl.gov 
     }

\begin{abstract}

\textbf{Objectives}: 
Lung cancer remains the leading cause of cancer-related mortality worldwide, and early diagnosis is critical for improving survival outcomes. However, early-stage malignancies often appear subtle on chest X-rays, posing a significant challenge for radiologists. This study explores Vision Transformers (ViTs) for predictive modeling of lung cancer one to two years before clinical diagnosis, advancing the frontier of proactive oncology. 

\textbf{Materials and Methods}:  
We analyzed approximately 259,000 chest X-rays from 91,000 imaging studies collected at the Jamaica Plains VA Hospital, Boston, MA. The dataset exhibits extreme class imbalance (approximately 1:150, cancer to non-cancer). To mitigate bias, we applied hybrid resampling strategies, combining under- and over-sampling with class-weighted loss optimization. Three ViTs configurations were evaluated: (a) trained from scratch, (b) ImageNet-pretrained, and (c) Corona (pneumonia)-pretrained models fine-tuned on the lung cancer dataset.

\textbf{Results}:
Transfer learning markedly improved performance, with pretrained models outperforming the scratch baseline by 6–10 percentage points in area under the curve (AUC) and approximately 10–12\% in balanced accuracy. The ImageNet-pretrained ViT achieved the most stable generalization (AUC $\approx$ 60–65\%, Balanced Accuracy $\approx$ 63–64\%), while Corona-pretrained models exhibited higher sensitivity but greater metric variance. Moderate resampling ratios (e.g., 1:1 or 1.5:2) balanced sensitivity and precision effectively, reducing computational time by up to 70\% without sacrificing accuracy.

\textbf{Conclusion}:
ViTs demonstrate strong potential for early lung cancer prediction from routine chest X-rays, particularly when leveraging transfer learning and balanced sampling. Although current performance remains below clinical deployment thresholds, these results establish a viable foundation for integrating ViTs into triage systems that flag high-risk patients for early intervention.

\end{abstract}

\blfootnote{Notice: Office of Science of the U.S. Department of Energy. This manuscript has been authored by UT-Battelle, LLC, under contract DE-AC05-00OR22725 with the US Department of Energy (DOE). The US government retains and the publisher, by accepting the article for publication, acknowledges that the US government retains a nonexclusive, paid-up, irrevocable, worldwide license to publish or reproduce the published form of this manuscript, or allow others to do so, for US government purposes. DOE will provide public access to these results of federally sponsored research in accordance with the DOE Public Access Plan (http://energy.gov/downloads/doe-public-access-plan). }



\section{Introduction}
Lung cancer remains a leading cause of cancer-related mortality worldwide, accounting for millions of deaths annually \cite{siegel2025cancer}. Early detection through low-dose computed tomography (CT) screening significantly reduces mortality, however screening uptake remains strikingly low, with fewer than 16\% of eligible individuals undergoing annual screening in the United States \cite{ALA2024}. As a result, most patients continue to be diagnosed in advanced stages, which underscores the need for complementary strategies that enable earlier risk identification from routine clinical imaging.

Recent advancements in artificial intelligence (AI) and deep learning have revolutionized medical imaging, offering new avenues for enhancing lung cancer diagnostics \cite{kaul2024ai}.One of the most critical challenges in lung cancer management is early identification of patients who are at the highest risk of developing lung cancer. Addressing this challenge would enable clinicians to design more precise therapeutic surveillance plans, thereby improving long-term patient outcomes. Existing methodologies lack the predictive capabilities to forecast cancer development years in advance. To address this gap, this study presents a model designed to classify chest radiographs according to whether a lung cancer diagnosis occurs within one or two years after imaging—thereby predicting disease onset within a defined interval. This approach provides a clinically meaningful framework for proactive oncology by identifying imaging patterns associated with future diagnoses.

Chest radiography, one of the most common imaging tests in healthcare, captures subtle pulmonary and mediastinal changes that may precede a cancer diagnosis. However, these early radiographic findings are often imperceptible to human observers and are not routinely used for predictive screening. Because chest radiography is routinely performed in clinical practice, incorporating AI-based risk prediction directly into chest x-ray (CXR) interpretation could enable opportunistic identification of high-risk individuals. This approach may help address the limited uptake of dedicated CT screening by leveraging existing imaging data without requiring additional patient visits \cite{raghu2022validation,jonas2021screening}. Conventional AI models based on convolutional neural networks (CNNs) have shown strong performance in medical imaging but remain limited in capturing global contextual patterns and rely heavily on large annotated datasets \cite{litjens2017survey,javed2024deep}. In contrast, Vision Transformers (ViTs) leverage self-attention mechanisms to model long-range dependencies and global features effectively \cite{dosovitskiy2022transformer,vaswani2024survey}. These attributes make ViTs particularly suited for the subtle and complex features inherent in early-stage lung cancer detection \cite{wang2022transformers,zhang2023biomedical}.

ViT have shown significant promise in medical vision tasks, including disease classification, segmentation, and anomaly detection. Their ability to process high-resolution images and leverage pretrained models has proven advantageous for biomedical applications, including cancer detection \cite{wang2022transformers,zhang2023biomedical}. For instance, recent studies highlight VIT superior performance in detecting abnormalities in chest X-rays, retinal images, and brain MRIs \cite{wang2022transformers}. Additionally, ViTs demonstrate robustness in learning from smaller datasets by leveraging transfer learning from large-scale, domain-agnostic pretrained models \cite{zhang2023biomedical}. In the context of lung cancer detection, traditional AI methods primarily focus on classification and segmentation tasks without extending into early prediction. By introducing ViTs into this domain, we aim to address this limitation and explore whether attention-based architectures can identify lung cancer at its earliest visible stages—one to two years before clinical diagnosis—through image-based prediction of disease onset within a defined interval.

In this study, we evaluate the feasibility of ViT–based models for detecting early radiographic signatures of lung cancer from routine chest X-rays. Using a large dataset of over 150,000 labeled chest radiographs from the Veterans Health Administration (VHA), we trained and fine-tuned pretrained ViT architectures to classify images according to whether a cancer diagnosis occurred within one or two years of imaging. We further assessed the effects of pretraining, class imbalance mitigation, and sampling strategies on model performance. By establishing a reproducible framework for image-based prediction of lung cancer onset within defined diagnostic intervals, this work aims to advance clinically interpretable, data-efficient AI approaches for early disease detection and risk stratification.

\section{Methods}
\subsection{Data description}
We included all chest X-ray obtained as part of routine care at the Boston VHA station 523 between 2010 and 2022. For each image, we computed binary indicators for lung cancer diagnoses documented in the VA cancer registry within one year and two years after the date of chest X-ray imaging.

To ensure unbiased evaluation, data were split by patient ID into non-overlapping training and testing cohorts, and the same testing set was used for all experiments.

\subsubsection{Preprocessing}

In our analysis, we employed a streamlined preprocessing pipeline to prepare the medical images for input into the ViT model. The pipeline consisted of four key steps: i) format conversion, ii) resizing and normalization, and iii) tensor transformation. 

\textit{First}, the medical images originally in DICOM\footnote{Digital Imaging and Communications in Medicine} format were converted to JPEG\footnote{Joint Photographic Experts Group} format to simplify processing and enable compatibility with the preprocessing tools. This conversion ensured that the images retained their quality and diagnostic information, while becoming more accessible for subsequent transformations.  

\textit{Second}, all input images were resized to a fixed dimension of $224 \times 224$ pixels, ensuring compatibility with the ViT's patch-based architecture. This resizing step preserved the overall spatial structure of the images while standardizing their dimensions for uniform processing. The pixel intensity values were then normalized to match the expected input distribution of the pretrained ViT model. Specifically, each pixel intensity $I_{\text{original}}$ was first rescaled to the range $[0, 1]$ using:

\[
I_{\text{rescaled}} = \frac{I_{\text{original}}}{I_{\text{max}}},
\]

where $I_{\text{max}}$ is the maximum possible pixel intensity (e.g. 255 for 8-bit images). The rescaled intensities were further standardized using the mean ($\mu$) and standard deviation ($\sigma$) of the ImageNet dataset:

\[
I_{\text{normalized}} = \frac{I_{\text{rescaled}} - \mu}{\sigma}.
\]

Here, the values of $\mu = [0.485, 0.456, 0.406]$ and $\sigma = [0.229, 0.224, 0.225]$ were applied uniformly across all three replicated channels to ensure compatibility with the pretrained model.

\textit{Third}, the images were converted into PyTorch tensors to facilitate efficient batch processing and integration with the ViT framework. This preprocessing pipeline, starting from DICOM conversion to tensor transformation, ensured that the images were appropriately formatted and scaled for optimal utilization of the ViT model's capabilities.

\subsection{Experimental setup}
For the experiments, we used a single NVIDIA A100 graphics processing unit (GPU) with 80 GB of memory on a DGX\footnote{Deep GPU Xceleration} node \cite{nvidia_a100_80gb}. The code was implemented in Python 3.9 \cite{python_org} and PyTorch 1.13 \cite{pytorch_org}, and training jobs were submitted using the SLURM\footnote{Simple Linux Utility for Resource Management} workload manager with the following configuration: 1 node, 1 GPU per node, 60 GB memory, and a wall time limit of 20 hours.

\subsection{Vision Transformer (ViT)}

\begin{figure}
    \centering
    \includegraphics[width=0.8\linewidth]{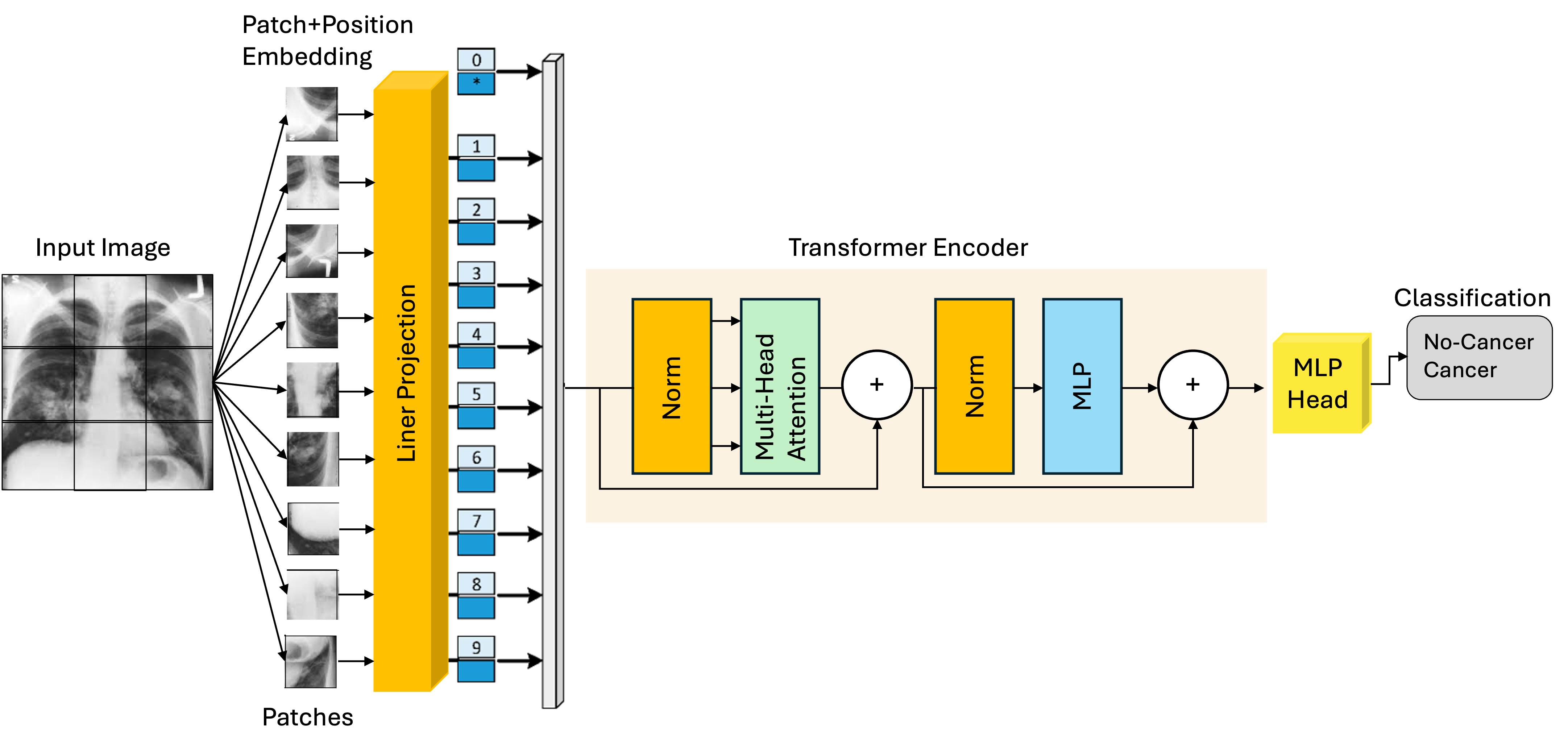}
    \caption{Vision Transformer (ViT) workflow for binary medical image classification. The model tokenizes the input image into patches, processes them through transformer encoders with self-attention, and outputs two classes: \textit{Cancer} and \textit{No-Cancer}.}
    \label{fig:ViT}
\end{figure}

The Vision Transformer (ViT) framework \cite{dosovitskiy2022transformer} adapts the transformer architecture originally developed for natural language processing to image data. Unlike convolutional neural networks (CNNs), which extract local features hierarchically, ViTs treat an image as a sequence of fixed-size patches (tokens), enabling modeling of long-range dependencies and global spatial context across the entire image (refer Figure \ref{fig:ViT}).

Image tokenization and embedding: Each image was divided into non-overlapping patches that were flattened and linearly projected into embeddings. A learnable classification token was prepended to aggregate global information, and positional encodings were added to retain spatial structure.

Transformer encoder: The sequence of patch embeddings was passed through multiple encoder layers composed of multi-head self-attention (MSA) and feed-forward networks (FFN) with residual connections and layer normalization to stabilize training.

Classification head: The final representation of the classification token was fed into a lightweight multilayer perceptron (MLP) to output the cancer versus no-cancer probabilities.

This formulation enables ViTs to learn both localized and global radiographic patterns, which are particularly important for subtle, early-stage lung cancer detection.
\subsection{Workflow methodology}
\begin{figure}
    \centering
    \includegraphics[width=0.8\linewidth]{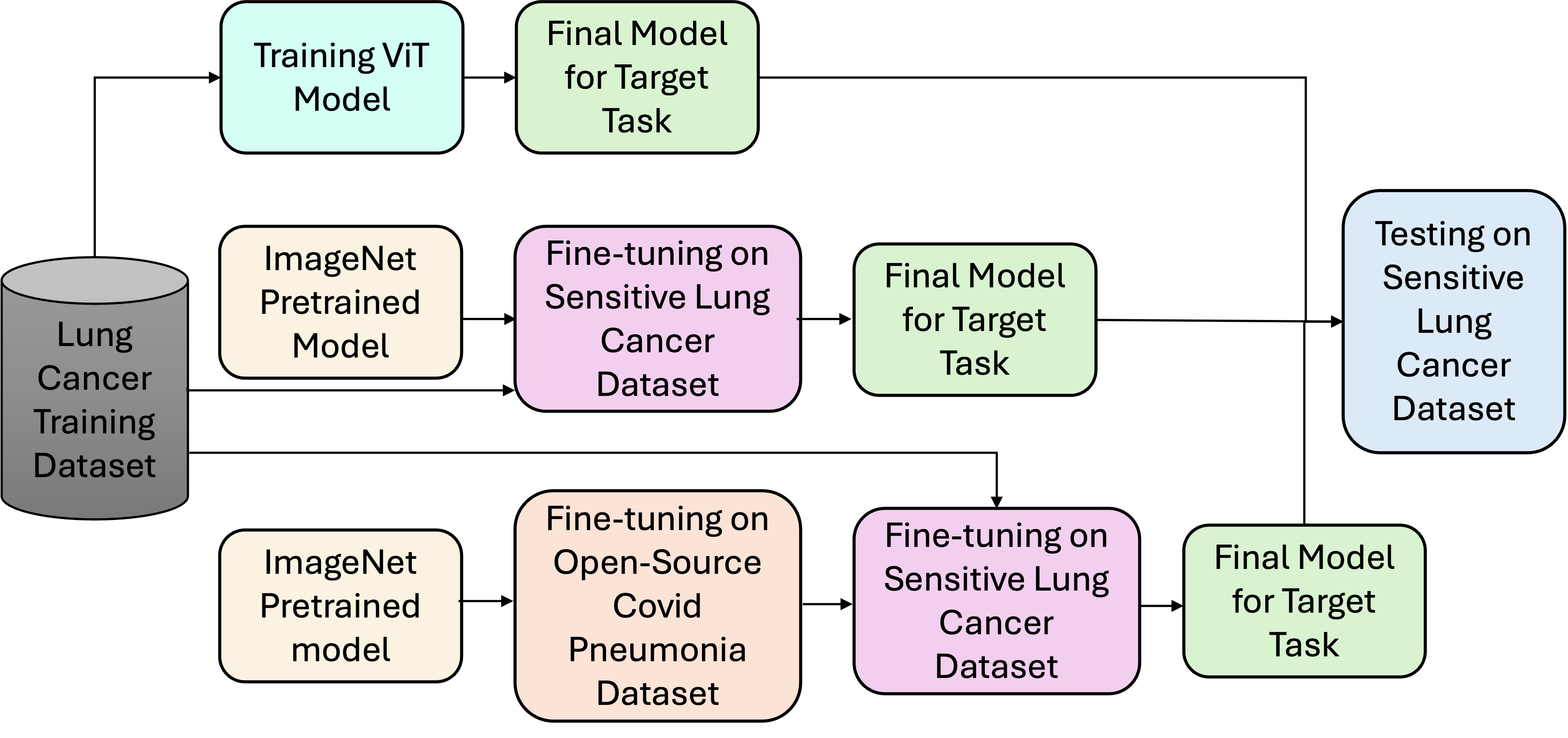}
    \caption{
    Overview of the experimental workflows for ViT fine-tuning on the lung cancer dataset. 
    The top pathway represents the baseline configuration trained directly on the sensitive dataset without pretraining. 
    The middle pathway illustrates the ImageNet-pretrained model fine-tuned on the lung cancer dataset. 
    The bottom pathway shows the two-stage fine-tuning strategy, where the model is first adapted using an open-source COVID pneumonia dataset before final fine-tuning on the sensitive lung cancer data. 
    All configurations are evaluated using the same testing dataset to ensure consistent performance comparison.
    }
    \label{fig:vit_workflow}
\end{figure}

For our study, we employed both non-pretrained and pretrained ViT configurations to evaluate model performance on the lung cancer imaging dataset. Figure~\ref{fig:vit_workflow} summarizes the overall architecture and experimental variants used.

\textbf{Non-Pretrained (From-Scratch) Model.}
We trained a ViT-Base model from scratch using the same architecture. The model uses a lightweight ViT with 4×4 patches, 2 transformer blocks, 2 attention heads, hidden dimension = 8, MLP ratio = 4, and 10 output classes. It was trained for 20 epochs using Adam optimizer (learning rate = 0.0003) \cite{kingma2014adam} with batch size = 32 and CrossEntropyLoss, applying weighted random sampling to handle class imbalance.

\textbf{Pretrained Model.} We fine-tuned two pretrained configurations of the ViT-Base-Patch16-224 architecture (via Hugging Face)~\cite{wolf2020transformers, googlevit2021}, each configured for binary classification. Both models were trained using the AdamW optimizer (\textit{lr} = $5\times10^{-5}$) and class-weighted CrossEntropyLoss to address severe class imbalance. Input images were resized to $224\times224$ RGB, normalized to a mean and standard deviation of 0.5, and loaded in batches of 32 using a WeightedRandomSampler based on class frequency. Each model was trained for 20 epochs and repeated three times where the data was randomly split between training and validation and no patient ID overlap between both. The best checkpoint selected according to balanced accuracy on the validation set.

Two pretrained configurations were evaluated:
\begin{itemize}
    \item \textbf{ViT-Imagenet:} Vision Base Model trained on Imagenet dataset. 86 million parameters, 12 transformer encoder layers, 768-dimensional embeddings, 12 self-attention heads, and a patch size of $16\times16$, initialized from the Hugging Face model \texttt{google/vit-base-patch16-224}~\cite{googlevit2021}.
    \item \textbf{ViT-Corona:} identical architecture, but ViT-Base model weights was further fine-tuned on the CoronaHack–ChestXRay dataset~\cite{chowdhury2020coronahack}, which comprises chest radiographs from healthy, pneumonia, and COVID-19 cases. This is test domain-specific transfer learning. 
\end{itemize}

\subsection{Handling the unbalanced classes}
To address the substantial class imbalance in our dataset ($>1:100$ ratio of cancer to non-cancer cases), we employed two complementary strategies: controlled sampling and loss reweighting. 

\textit{Controlled sampling} involved systematically adjusting the representation of each class to achieve more balanced training data. In the \textit{undersampling} approach, we selectively reduced the number of samples from the majority class to achieve predefined class ratios ranging from perfectly balanced (1:1) to moderately imbalanced (1:4). This controlled reduction prevented excessive data loss while ensuring that both classes contributed comparably during training. 

In contrast, the \textit{oversampling} approach focused on increasing the representation of the minority class by replicating or augmenting its samples to reach moderate ratios between 1:1 and 1.5:2. This strategy allowed us to evaluate the benefits of enhanced minority representation without risking overfitting due to excessive duplication. Across both sampling schemes, we conducted systematic experiments to identify configurations that offered the best trade-off between model robustness and computational efficiency. 

To further mitigate imbalance effects, we applied \textit{loss reweighting} during model optimization. Each class was assigned a weight inversely proportional to its frequency in the training data, so that errors in underrepresented classes contributed more strongly to the overall loss. This ensured that the model maintained balanced attention to all classes and avoided bias toward the majority class.

\subsection{Experiments and Analysis}

We created patient-level, disjoint splits to avoid leakage (no patient appears in more than one split). A fixed hold-out test set was kept identical across all experiments, and model development used the training/validation portions only. Each experiment was repeated three times with different random seeds; unless otherwise stated, we selected the best checkpoint by validation balanced accuracy and reported the mean and standard deviation.
We conducted a comprehensive evaluation of ViT models fine-tuned for lung cancer detection under various class ratios and pretraining sources. Each configuration was evaluated using the VA CXR dataset for two consecutive years. The experiments systematically assessed the impact of pretraining (ImageNet vs. Corona), sampling strategies (under- and over-sampling), and class imbalance on model performance. Each experiment was repeated three times, and mean values across ten-fold cross-validation are reported with standard deviations. Evaluation metrics included AUC and Balanced Accuracy.

\subsection{Metrics}

AUC (Area Under the Curve) and balanced accuracy were chosen as evaluation metrics because they effectively address the challenges posed by imbalanced datasets, such as those encountered in medical imaging for lung cancer diagnosis. 

\textit{AUC} measures the model's ability to distinguish between classes by evaluating its performance across all classification thresholds, providing a robust assessment of the trade-off between sensitivity or True Positive Rate (TPR) and specificity or True Negative Rate (TNR). The AUC provides a single scalar value summarizing the model's discriminative ability. Formally, AUC is computed as:

\[
\text{AUC} = \int_0^1 \text{TPR}(t) \, d\text{FPR}(t)
\]

where $\text{TPR}$ is the true positive rate and $\text{FPR}$ is the false positive rate, both evaluated over all thresholds $t$. AUC is particularly valuable in imbalanced settings, as it captures the overall separability of the classes without being biased by class distribution.

\textit{Balanced accuracy}, on the other hand, accounts for the unequal representation of classes by averaging the recall (sensitivity) for each class. It is defined as:

\[
\text{Balanced Accuracy} = \frac{1}{2} \left( \frac{\text{TP}}{\text{TP} + \text{FN}} + \frac{\text{TN}}{\text{TN} + \text{FP}} \right)
\]
This ensures that performance on minority samples is not masked by class imbalance.

\textit{Execution time} was measured as the elapsed wall-clock duration between training start time and testing end time.  

\section{Results}
\subsection{Dataset}
The VA-CXR from the Boston VHA station consisted of 259,361 chest X-rays associated with 91,020 imaging studies from 35,771 patients. Each study included multiple view positions—primarily posteroanterior (PA) and lateral (LL) views, with fewer anteroposterior (AP) images—reflecting typical clinical imaging protocols (see Table \ref{tab:stats}).

\begin{table}[h!]
\centering
\caption{Summary statistics for Year 1 and Year 2 datasets. It shows the number of samples per class, study and patient ID, and different view positions such as LL, PA, and AP.}
\label{tab:stats}
\begin{tabular}{|l|c|c|}
\hline
 & \textbf{Year 1 Dataset} & \textbf{Year 2 Dataset} \\ \hline
\# Number of Patients & 28,637 & 28,281 \\ \hline
\# Number of Studies & 63,818 & 63,322 \\ \hline
\# Chest X-ray Images & 169,001 & 167,684 \\ \hline
\multicolumn{3}{|l|}{\textit{Category-wise \# Images:}} \\ \hdashline 
\quad Class 0: No-cancer diagnosed & 167,936 & 166,124 \\ 
\quad \quad \textit{Viewposition-wise} & & \\ 
\quad \quad \quad Left Lateral (LL) & 77,088 & 76,243 \\ 
\quad \quad \quad Posteroanterior (PA)& 70,489 & 69,679 \\ 
\quad \quad \quad Anteroposterior (AP) & 7,688 & 7,616 \\ \hdashline
\quad Class 1: Cancer diagnosed & 1,065 & 1,560 \\ 
\quad \quad \textit{Viewposition-wise} & & \\ 
\quad \quad \quad Left Lateral (LL) & 483 & 722 \\ 
\quad \quad \quad Posteroanterior (PA) & 480 & 714 \\ 
\quad \quad \quad Anteroposterior (AP) & 35 & 42 \\ \hline
\end{tabular}
\end{table}

The dataset comparison across the two years reveals a consistent and well-structured collection of radiographic studies, with minimal variation in patient and study counts, indicating stable data acquisition practices. Both years exhibit a pronounced class imbalance, with significantly more no-cancer (Class 0) than cancer-diagnosed (Class 1) cases, though Year 2 shows an increase in positive samples (from 1,065 to 1,560). The distribution of imaging view positions—primarily PA and LL, with fewer AP images previously reported in [self-citation redacted] 
— remains proportionally similar across years, reflecting consistent clinical imaging protocols. The higher proportion of PA and LL views compared to AP views indicates that most scans were obtained under standard, upright conditions, which may bias the model toward recognizing patterns typical of ambulatory patients rather than those imaged in supine or critical care settings. Overall, the Year 2 dataset maintains continuity with Year 1 while modestly enriching the minority cancer class, supporting more balanced and generalizable model training.
All subsequent experiments used identical patient-level splits, ensuring no overlap between training and testing sets.

\subsection{Pretrained model evaluation across sampling techniques}

Tables \ref{tab:ratio-model-auc} and \ref{tab:ratio-model-balacc} summarize the average AUC and Balanced Accuracy scores across different sampling ratios. The results show consistent trends across years and pretraining sources.

\begin{table}[h!]
\centering
\caption{AUC Scores (est. mean ± std) across different class ratios and pretrained models. U = undersampling and O = oversampling.}
\begin{tabular}{|p{1.5cm}|p{2cm}|c|c||c|c|}\hline
 \multirow{2}{*}{\textbf{Sampling}}& \multirow{2}{*}{\textbf{Class Ratio}}& \multicolumn{2}{|c||}{\textbf{Year 1}}& \multicolumn{2}{|c|}{\textbf{Year 2}}\\\cline{3-6}
 & & \textbf{VIT-ImageNet}& \textbf{VIT-Corona}& \textbf{VIT-ImageNet}&\textbf{VIT-Corona}\\ \hline
- & Original Dataset & 59.24 ± 0.85  &59.70 ± 1.43 
& 57.78 ± 1.12 & 58.83 ± 1.05 \\ \hline
\multirow{3}{1.5cm}{Under sampling} & U: 1:1 & 64.07 ± 1.67  &61.51 ± 1.10 
& 60.79 ± 0.94 & 57.15 ± 1.42 \\ \cline{2-6}
&U: 1:2 & 58.18 ± 1.32  &58.23 ± 0.93 
& 58.29 ± 1.25 & 58.18 ± 1.06 \\ \cline{2-6}
&U: 1:3 & 56.65 ± 1.21  &55.98 ± 0.87 
& 58.26 ± 1.08 & 55.84 ± 1.10 \\ \hline
\multirow{3}{1.5cm}{Over sampling}&O: 1.5:1.5 & 60.04 ± 1.54  &61.63 ± 1.11 
& 58.25 ± 0.95 & 57.27 ± 1.26 \\ \cline{2-6}
&O: 1.5:2 & 58.24 ± 1.38  &58.28 ± 0.97 
& 58.59 ± 1.20 & 60.21 ± 1.23 \\ \cline{2-6}
&O: 1.5:3 & 58.28 ± 1.45  &58.56 ± 1.05 & 50.47 ± 1.75 & 60.32 ± 1.30 \\ \hline
\end{tabular}
\label{tab:ratio-model-auc}
\end{table}

\begin{table}[ht]
\centering
\caption{Balanced Accuracy (mean ± std) across different class ratios and pretrained models. U = undersampling and O = oversampling.}
\begin{tabular}{|p{1.5cm}|p{2cm}|c|c||c|c|}\hline
  \multirow{2}{*}{\textbf{Sampling}}&\multirow{2}{*}{\textbf{Class Ratio}}& \multicolumn{2}{|c|}{\textbf{Year 1}}& \multicolumn{2}{|c|}{\textbf{Year 2}}\\\cline{3-6}
 & & \textbf{VIT-ImageNet}&  \textbf{VIT-Corona}&\textbf{VIT-ImageNet}& \textbf{VIT-Corona}\\ \hline
- &Original Dataset & 52.23 ± 1.17 &  62.27 ± 3.44 
&61.30 ± 1.38 & 59.02 ± 1.55 \\ \hline
 \multirow{3}{1.5cm}{Under sampling} &U: 1:1      & 64.7 ± 1.52 &  59.73 ± 2.80 
&62.47 ± 1.15 & 59.15 ± 1.67 \\ \cline{2-6}
 &U: 1:2      & 61.43 ± 1.57 &  60.29 ± 1.92 
&63.39 ± 0.60 & 61.90 ± 1.32 \\ \cline{2-6}
 &U: 1:3      & 64.43 ± 1.26 &  59.07 ± 2.55 
&62.13 ± 1.19 & 59.63 ± 2.19 \\ \hline
 \multirow{3}{1.5cm}{Over sampling}&O: 1.5:1.5  & 65.00 ± 2.00 &  63.60 ± 1.17 
&63.13 ± 0.23 & 61.37 ± 0.47 \\ \cline{2-6}
 &O: 1.5:2    & 60.47 ± 1.66 &  61.47 ± 1.63 
&61.27 ± 1.47 & 62.80 ± 1.29 \\ \cline{2-6}
 &O: 1.5:3    & 59.13 ± 2.58 &  60.47 ± 2.01 &59.57 ± 4.49 & 60.80 ± 0.90 \\ \hline

\end{tabular}
\label{tab:ratio-model-balacc}
\end{table}

Across all configurations, models initialized from ImageNet dataset is mostly outperform those initialized from the Corona dataset in both Balanced Accuracy and AUC, confirming the advantage of pretraining on large-scale, diverse datasets. 
On the other hand, regarding balanced accuracy.... For the whole dataset, ImageNet-based fine-tuning achieved balanced accuracies of 52.23–61.30\%, compared to 59.02–62.27\% for Corona-initialized models. 
When compared to the non-pretrained baseline, which achieved only 49.77–51.75\% balanced accuracy, both pretrained models demonstrate substantial improvements, confirming the necessity of transfer learning for small medical datasets. 

The gain from pretraining is particularly visible in AUC scores, where ImageNet-based fine-tuning surpasses the non-pretrained model by 6–10 percentage points. These improvements are attributed to the transfer of generalized visual representations that enable more effective feature extraction for lung cancer.

\subsection{Visual Performance Comparison}

To better characterize metric-level behavior, radar plots were generated to visualize the joint distribution of Balanced Accuracy, Sensitivity, Specificity, Precision, and F1-score across configurations. 

\begin{figure}[htbp]
    \centering
    \begin{subfigure}[t]{0.48\linewidth}
        \centering
        \includegraphics[width=\linewidth]{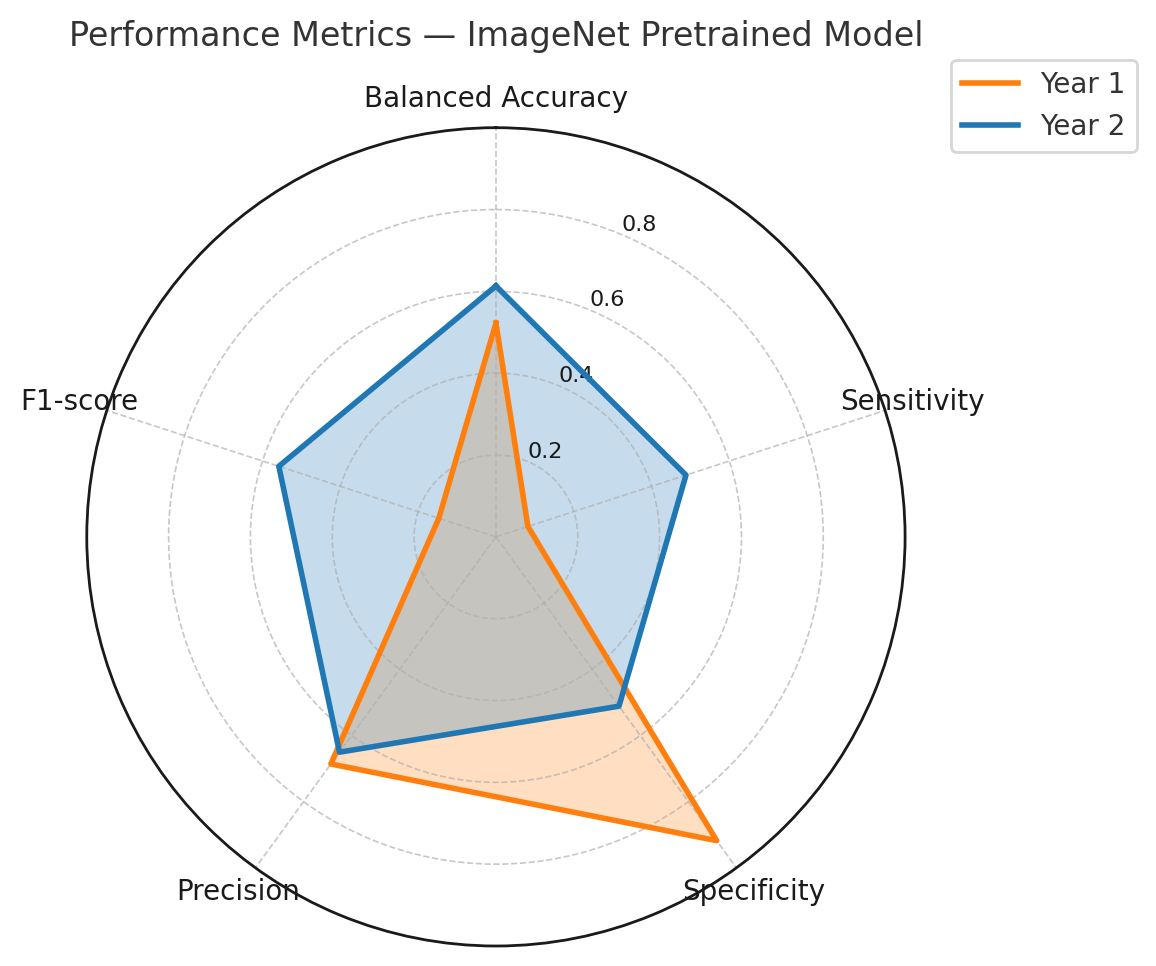}
        \caption{ImageNet-pretrained model performance across metrics for Year 1 and Year 2.}
        \label{fig:radar-imagenet}
    \end{subfigure}
    \hfill
    \begin{subfigure}[t]{0.48\linewidth}
        \centering
        \includegraphics[width=\linewidth]{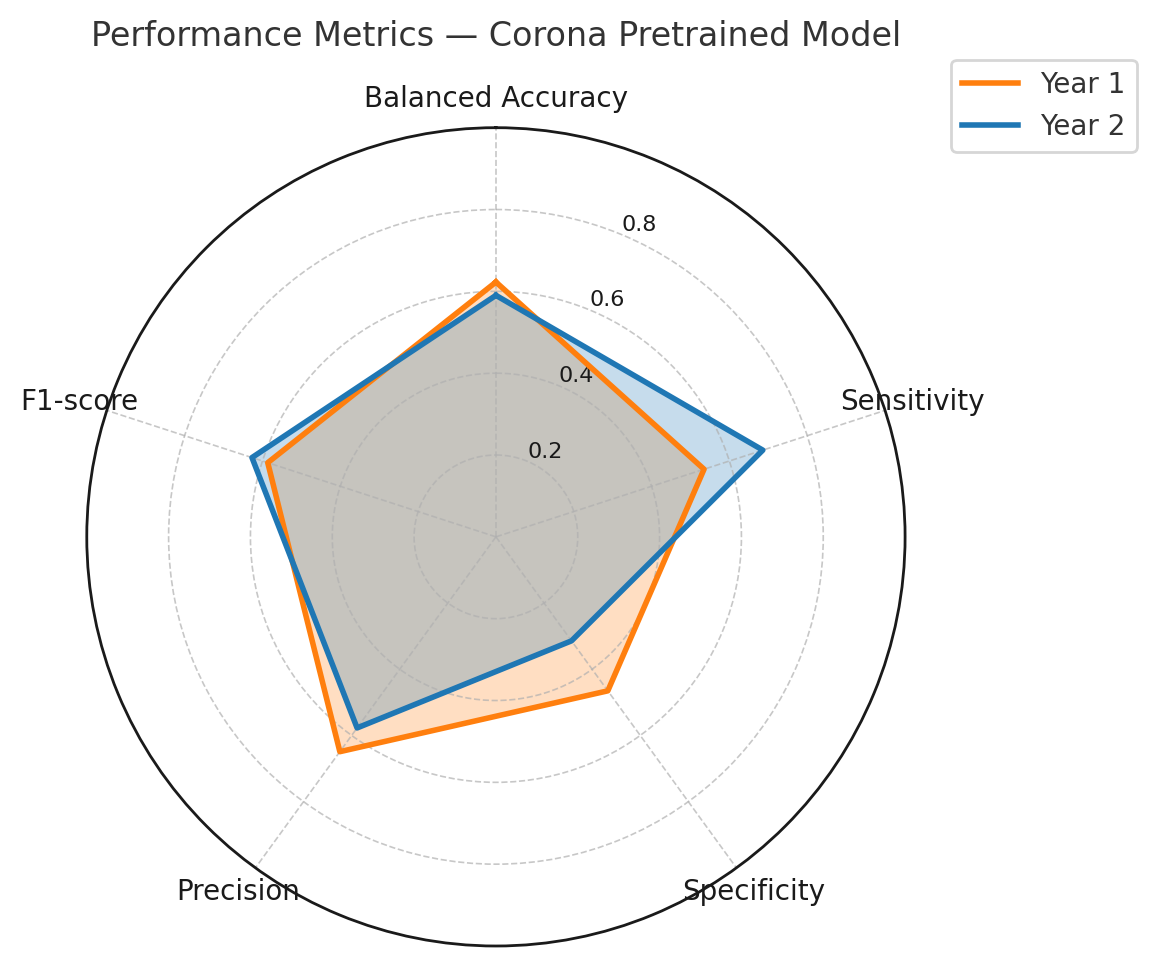}
        \caption{Corona-pretrained model performance across metrics for Year 1 and Year 2.}
        \label{fig:radar-corona}
    \end{subfigure}
    \caption{Side-by-side radar plots comparing the evolution of model performance from Year 1 to Year 2 using two pretraining strategies. Both models are evaluated on the same five metrics: Balanced Accuracy, Sensitivity, Specificity, Precision, and F1-score.}
    \label{fig:radar-comparison}
\end{figure}

Figure \ref{fig:radar-comparison} compares the evolution of ViT model performance between Year 1 and Year 2 for two pretraining sources—ImageNet and Corona—evaluated across Balanced Accuracy, Sensitivity, Specificity, Precision, and F1-score.
The ImageNet-pretrained model shows substantial gains in Sensitivity ($\approx$ 0.08 → 0.49) and F1-score ($\approx$ 0.15 → 0.56), indicating improved detection of cancer-positive cases after extended fine-tuning. Specificity declines from $\approx$ 0.92 to 0.51, suggesting that the model becomes more sensitive but slightly less conservative when predicting negatives. Overall Balanced Accuracy remains stable, confirming that recall improvements do not severely distort global performance.
The Corona-pretrained model exhibits a more balanced pattern of growth, with Sensitivity ($\approx$ 0.53 → 0.69) and F1-score ($\approx$ 0.59 → 0.63) improving modestly while Balanced Accuracy stays near 0.60. Despite minor reductions in Specificity, the resulting polygons are more symmetric, reflecting steadier trade-offs between recall and precision.
Quantitatively, the ImageNet-initialized model achieves a larger normalized radar-area score (0.83 ± 0.04) across all five metrics, compared with 0.72 ± 0.06 for the Corona-initialized model—evidence of stronger and more uniform generalization. The narrower area and higher metric variance observed for Corona models indicate reduced stability and partial catastrophic forgetting of general visual features caused by intermediate domain-specific pretraining. By contrast, direct ImageNet fine-tuning preserves broad representational diversity, enabling more reliable transfer and better metric balance across evaluation axes.

\begin{figure}[htbp]
    \centering
    \includegraphics[width=0.95\linewidth]{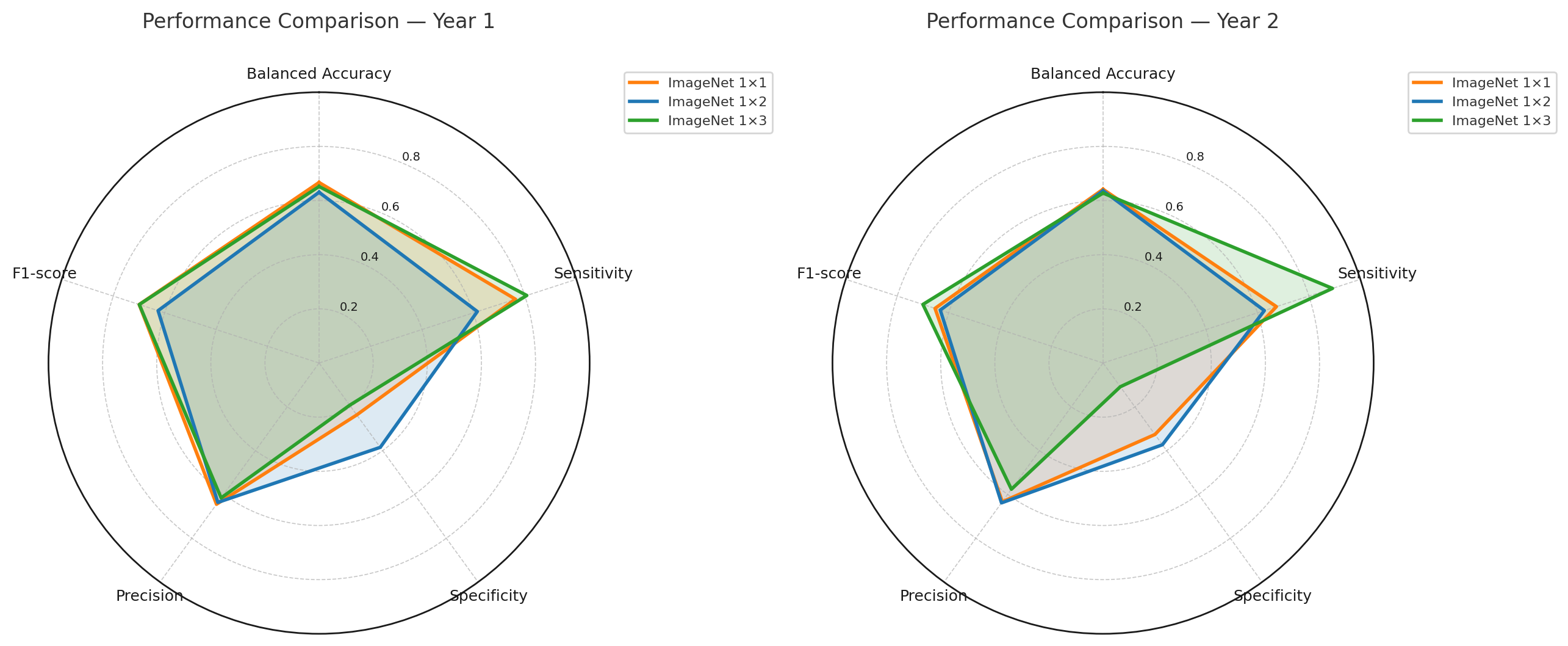}
    \caption{Side-by-side comparison of ImageNet-pretrained Vision Transformer models across Year 1 (left) and Year 2 (right). 
    Each polygon represents a fine-tuning configuration (1×1, 1×2, 1×3), evaluated on five metrics: Balanced Accuracy, Sensitivity, Specificity, Precision, and F1-score.}
    \label{fig:radar-imagenet-year1-year2}
\end{figure}

Figure \ref{fig:radar-imagenet-year1-year2} presents detailed metric profiles for ImageNet-pretrained ViT models fine-tuned on the lung-cancer dataset under three under sampling configurations (1×1, 1×2, 1×3)

Each polygon corresponds to one configuration, with axes representing Balanced Accuracy, Sensitivity, Specificity, Precision, and F1-score for Year 1 (left) and Year 2 (right).
In Year 1, the 1×3 configuration forms the broadest polygon, achieving high Sensitivity ($\approx$ 0.81) and F1-score ($\approx$ 0.70) but low Specificity ($\approx$ 0.19). This recall-dominant shape reflects aggressive positive detection and elevated false-positive rates. As the sampling ratio decreases (1×3 → 1×1), the polygons contract and become more uniform, signifying improved precision and stability at the cost of slightly lower recall.
By Year 2, all three configurations converge around Balanced Accuracy $\approx$ 0.63–0.64, showing that extended training promotes consistent cross-metric performance. The 1×2 configuration yields the most evenly distributed polygon (F1 $\approx$ 0.63; Specificity $\approx$ 0.37), balancing recall and precision while maintaining general robustness. This pattern mirrors the higher radar-area score of ImageNet-based models in Figure 2, reaffirming that moderate fine-tuning ratios maximize generalization efficiency and reduce metric variance. Excessive oversampling enhances recall but narrows Specificity, whereas smaller ratios produce conservative yet reproducible performance suitable for clinical deployment.

\begin{figure}[htbp]
    \centering
    \includegraphics[width=0.95\linewidth]{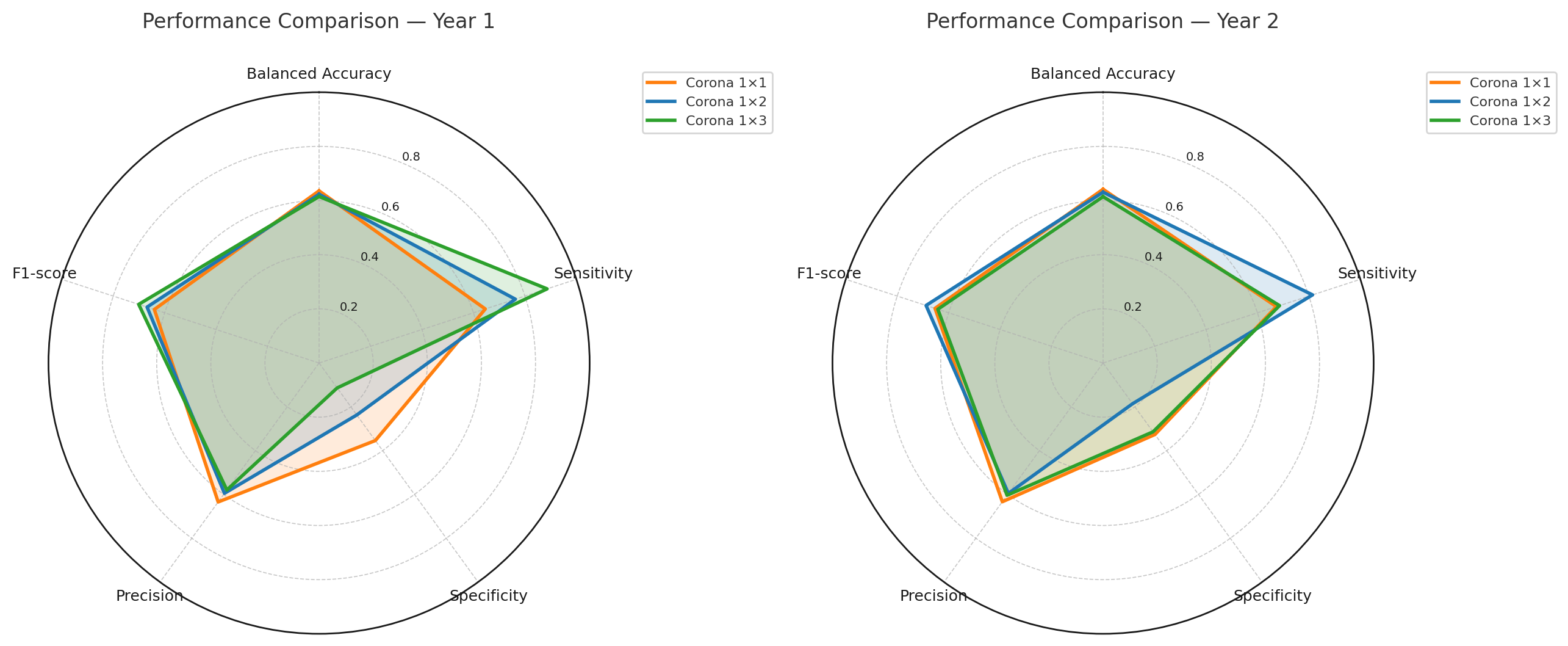}
    \caption{Side-by-side comparison of Corona-pretrained Vision Transformer models across Year 1 (left) and Year 2 (right). 
    Each polygon represents a fine-tuning configuration (1×1, 1×2, 1×3), evaluated across five metrics: Balanced Accuracy, Sensitivity, Specificity, Precision, and F1-score.}
    \label{fig:radar-corona-year1-year2}
\end{figure}

Figure \ref{fig:radar-corona-year1-year2} shows the corresponding results for Corona-pretrained ViT models, fine-tuned using the same three configurations (1×1, 1×2, 1×3) and evaluated across identical metrics.

During Year 1, the 1×3 configuration yields the highest Sensitivity ($\approx$ 0.89) and F1-score ($\approx$ 0.70) but exhibits marked Specificity reduction, indicating a recall-oriented decision boundary. The 1×1 configuration forms a smaller, more compact polygon, representing balanced but conservative predictions.
By Year 2, the polygons across all configurations become smoother and more symmetric, highlighting improved stability and convergence. The 1×2 configuration emerges as the most balanced, achieving Sensitivity $\approx$ 0.81, F1 $\approx$ 0.69, and Balanced Accuracy $\approx$ 0.63. Meanwhile, the 1×3 model retains high recall but reduces its metric asymmetry, implying gradual correction of domain-specific overfitting.
Relative to ImageNet-based models (Figure 4), these Corona-initialized models occupy smaller radar-areas and exhibit higher metric variance, consistent with the 0.72 ± 0.06 normalized radar-area found in Figure 2. This confirms that intermediate domain adaptation introduces instability and partial forgetting of general representations, while additional fine-tuning partially restores equilibrium across recall-precision trade-offs. Nevertheless, their steadier Year 2 evolution demonstrates that extended optimization can mitigate early over-specialization and enhance transferability to the lung-cancer task.

\subsection{Execution Time}
Full-dataset training and testing required approximately 16 to 30 hours, whereas under-sampling and over-sampling configurations reduced the training time to 14–18 hours per experiment—corresponding to a 25–70\% reduction in runtime without noticeable performance degradation. This improvement demonstrates that appropriately chosen sampling ratios (e.g., under-sampling 1:1 or over-sampling 1.5:2) can preserve model robustness while substantially lowering computational costs. Analysis of training durations further illustrates the trade-offs in efficiency across different sampling strategies.

\section{Discussion}

Our findings highlight several critical aspects in the development of reliable deep-learning models for the diagnosis of lung cancer from chest radiographs. Together, the results indicate that routine chest radiographs contain weak but detectable imaging signals associated with lung cancer up to one to two years before clinical diagnosis \cite{lu2020deep}. However, the signal from the chest radiograph remains insufficient for prediction alone and performance plateaus below the thresholds required for diagnostic use \cite{raghu2022validation, mikhael2023sybil,ardila2019end}. These findings suggest that, while early predictive information exists in CXRs, its clinical utility depends on contextual integration rather than isolated image-based inference \cite{patz2014overdiagnosis}.

Given these limitations in signal strength, improving model performance requires approaches that can leverage external knowledge beyond the dataset itself. First, the experiments confirm that transfer learning from pretrained models substantially improves performance compared to training from scratch. Models initialized with pretrained weights consistently surpassed the untrained baseline by 6–10 percentage points in AUC and by approximately 10–12\% in balanced accuracy. This improvement demonstrates that pretrained models capture generalized visual features—such as edges, textures, and anatomical structures—that are transferable to downstream medical tasks, even when the target dataset is relatively small or imbalanced \cite{dosovitskiy2022transformer, raghu2019transfusion, tajbakhsh2016convolutional}.

Second, the extreme class imbalance of the VA CXR dataset ($\approx$1:150 cancer-to-non-cancer ratio) proved to be a decisive factor influencing model behavior. Without correction, models tended to overfit to the majority class, yielding artificially high specificity but extremely low sensitivity. Balancing through under-sampling or moderate over-sampling (e.g., 1:1 or 1.5:2 ratios) improved sensitivity and F1-score while maintaining comparable precision. However, heavy resampling (e.g., 1:3) led to instability and degraded generalization, emphasizing the need for careful control of sampling strategies in medical imaging tasks with rare positive cases \cite{saito2015precision,buda2018systematic}.

Third, the comparison between ImageNet and Corona pretraining underscores a nuanced trade-off between generalization and domain adaptation. ImageNet-pretrained models achieved higher overall stability and balanced accuracy, reflecting the advantages of large-scale, heterogeneous training data that encourage general-purpose feature learning. In contrast, Corona-pretrained models—despite their domain proximity to lung pathologies—showed more variability across metrics and occasionally lower precision. This result suggests that while domain-specific pretraining captures relevant radiographic features, it may also introduce narrow feature biases that limit transferability to new disease categories such as lung cancer. In essence, large-scale generic pretraining provides robust generalization, whereas domain-specific pretraining offers contextual alignment at the risk of overfitting to the source pathology distribution \cite{kirkpatrick2017overcoming,chowdhury2020can}.

Interestingly, when the same ViT architecture was applied to X-ray classification of COVID pneumonia, convergence was rapid, with accuracy exceeded 90\% after only a few training steps. However, when transferred to the lung cancer detection task under the same configuration, performance plateaued around 60–65\% balanced accuracy. This discrepancy arises from intrinsic differences in radiographic presentation: COVID pneumonia manifests as spatially diffuse opacities, whereas early-stage lung cancer lesions are subtle, localized, and often confounded by anatomical variation \cite{chowdhury2020can}.
Consequently, detecting cancer demands higher spatial resolution and stronger attention to fine-grained features beyond single-view radiographs. 

From a clinical standpoint, these results have several implications. Although transfer learning accelerates model development and improves detection sensitivity, the absolute performance levels remain below clinical acceptability thresholds for diagnostic use. Chest X-rays are frequently used as a first-line screening tool due to their accessibility and low cost, but their inherent limitations in resolving small or early lesions constrain the achievable accuracy of AI models trained solely on this modality. Future deployment in clinical workflows should therefore treat these models as triage or assistive systems rather than definitive diagnostic tools, complementing radiologists by flagging high-risk cases for further review. In practice, such a system could be integrated into routine CXR interpretation pipelines, flagging high-risk patients for follow-up imaging or low dose computed tomography (LDCT) referral rather than providing definitive diagnoses. Given the low prevalence of lung cancer in routine CXRs, even moderate false-positive rates could result in a substantial downstream workload. Future work must therefore prioritize precision and calibration alongside sensitivity to ensure clinical feasibility \cite{bach2012benefits}.

Finally, computational efficiency also plays an important practical role. Training on the full VA CXR dataset required two days per experiment, whereas balanced subsampling configurations reduced training time by up to 70\% to XYZ without significant accuracy loss. This efficiency gain is critical for real-world hospital settings, where continuous model retraining or domain adaptation must operate within strict resource and time constraints.

In summary, the study demonstrates that (1) pretrained models outperform untrained baselines, (2) moderate resampling effectively mitigates data imbalance, (3) ImageNet pretraining yields more stable and generalizable models than disease-specific pretraining, and (4) the complexity of lung cancer detection from X-rays requires careful tuning of model capacity, data balance, and transfer-learning strategies to ensure clinically meaningful and computationally efficient outcomes. Collectively, these findings support a shift from purely diagnostic objectives toward risk stratification frameworks that leverage routinely acquired imaging to identify patients who may benefit from earlier intervention or intensified surveillance.

This study represents an initial step toward image-based early risk stratification for lung cancer, and several opportunities for future refinement remain. First, while the dataset was derived from a single VHA site, this provides a well-controlled clinical environment that can serve as a foundation for future multi-institutional validation across more diverse populations. Second, cancer outcomes were defined using registry diagnosis dates rather than radiographic confirmation of disease onset, motivating future work that incorporates longitudinal imaging trajectories to better characterize temporal disease evolution. Third, although external validation was not performed in this study, the standardized modeling framework enables straightforward extension to independent cohorts. Fourthly, view-position heterogeneity and temporal relationships between serial chest radiographs were not explicitly modeled, presenting clear opportunities for multi-view and longitudinal architectures. Finally, model calibration and explainability were beyond the scope of this analysis but represent important next steps toward clinical integration and prospective evaluation.

\section{Conclusion}
This work evaluated ViT models for lung cancer detection from chest X-rays, focusing on the effects of pretrained initialization, data imbalance, and domain-specific transfer learning. Both ImageNet- and Corona-pretrained models outperformed untrained baselines, underscoring the value of transfer learning for limited medical datasets. ImageNet pretraining yielded more stable and balanced performance across metrics, while Corona pretraining—though domain-relevant—showed higher sensitivity but greater variability. Moderate sampling strategies mitigated imbalance effectively, improving recall without major accuracy loss. Despite these gains, overall performance remains below clinical thresholds, reflecting the difficulty of identifying subtle cancer lesions from X-rays. Future work will explore multimodal and federated approaches to enhance diagnostic reliability and computational efficiency in real-world healthcare settings.

\section{Acknowledgments}
This manuscript has been co-authored by UT-Battelle, LLC under Contract No. DE-AC05-00OR22725 with the U.S. Department of Energy. The publisher, by accepting the article for publication, acknowledges that the U.S. Government retains a non-exclusive, paid up, irrevocable, world-wide license to publish or reproduce the published form of the manuscript, or allow others to do so, for U.S. Government purposes. The DOE will provide public access to these results in accordance with the DOE Public Access Plan (http://energy.gov/downloads/doe-public-access-plan).
This research is supported by the Million Veteran Program, Office of Research and Development, Veterans Health Administration, and was supported by the MVP-000 award. This publication does not represent the views of the Department of Veteran Affairs or the United States Government.
Our team would like to acknowledge the support of the Knowledge Discovery Infrastructure team. 

\printbibliography

%
%

\clearpage

\onecolumn
\setcounter{figure}{0} \renewcommand{\thefigure}{A.\arabic{figure}}
\setcounter{table}{0} \renewcommand{\thetable}{A.\arabic{table}}


\end{document}